\documentclass[10pt,twocolumn]{article}
\usepackage{graphicx}
\usepackage{amsmath}  
\usepackage{hyperref} 
\usepackage[affil-it]{authblk} 
\usepackage{amssymb}
\usepackage{booktabs}
\usepackage{comment}
\usepackage{multirow}
\usepackage{makecell}
\usepackage{cleveref}
\usepackage{float}
\usepackage[utf8]{inputenc}
\usepackage[margin=0.55in]{geometry}

\renewcommand{\abstractname}{\large\textbf{Abstract}}
\renewenvironment{abstract}
  {\center\Large\textbf{\abstractname}\par\normalsize\par\noindent}
  {\vspace{0.5cm}}
\makeatletter
\renewcommand\title[1]{\gdef\@title{\Large\bfseries #1}} 
\makeatother

\title{Weakly Supervised Contrastive Learning for Histopathology Patch Embeddings}

\begin{document}

\author{
    \upshape{Bodong Zhang}\textsuperscript{\dag1,2 [0000-0001-9815-0303]} \quad {Xiwen Li}\textsuperscript{2 [0009-0005-6139-0668]} \quad {Hamid Manoochehri}\textsuperscript{1,2 [0009-0005-0478-7925]} \quad {Xiaoya Tang}\textsuperscript{2 [0009-0002-5638-320X]} \quad {Deepika Sirohi}\textsuperscript{\ddag3 [0000-0002-0848-4172]} \quad {Beatrice S. Knudsen}\textsuperscript{3 [0000-0002-7589-7591]} \quad {Tolga Tasdizen}\textsuperscript{1,2 [0000-0001-6574-0366]}\\\vspace{5pt}
    {\fontsize{10.5pt}{12pt}\selectfont \textsuperscript{1}Department of Electrical and Computer Engineering, University of Utah, Salt Lake City, UT, USA\\ \textsuperscript{2}Scientific Computing and Imaging Institute, University of Utah, Salt Lake City, UT, USA\\ \textsuperscript{3}Department of Pathology, University of Utah, Salt Lake City, UT, USA
    % \\\upshape\texttt{Key words: Weakly supervised learning, Contrastive learning, Multiple instance learning, Multi-task learning, Digital histopathology whole slide images}
    \\\upshape{Keywords: Weakly supervised learning, Contrastive learning, Multiple instance learning, Multi-task learning, Digital histopathology whole slide images}
    %\texttt{hamid.manoochehri@utah.edu},  %\\\upshape\texttt{beatrice.knudsen@pathology.utah.edu}, \texttt{tolga.tasdizen@utah.edu}
    }}

\date{}

\twocolumn[
  \begin{@twocolumnfalse}
    \maketitle
    \begin{abstract}
    %\raggedright
    Digital histopathology whole slide images (WSIs) provide gigapixel-scale high-resolution images that are highly useful for disease diagnosis. However, digital histopathology image analysis faces significant challenges due to the limited training labels, since manually annotating specific regions or small patches cropped from large WSIs requires substantial time and effort. Weakly supervised multiple instance learning (MIL) offers a practical and efficient solution by requiring only bag-level (slide-level) labels, while each bag typically contains multiple instances (patches). Most MIL methods directly use frozen image patch features generated by various image encoders as inputs and primarily focus on feature aggregation. However, feature representation learning for encoder pretraining in MIL settings has largely been neglected. In our work, we propose a novel feature representation learning framework called weakly supervised contrastive learning (WeakSupCon) that incorporates bag-level label information during training. Our method does not rely on instance-level pseudo-labeling, yet it effectively separates patches with different labels in the feature space. Experimental results demonstrate that the image features generated by our WeakSupCon method lead to improved downstream MIL performance compared to self-supervised contrastive learning approaches in three datasets. Our related code is available at \url{github.com/BzhangURU/Paper_WeakSupCon_for_MIL}
    \end{abstract}
    \vspace{0.5cm}
  \end{@twocolumnfalse}
]

%% FOOTER (equal contribution and corresponding author)
\renewcommand{\thefootnote}{\fnsymbol{footnote}}
%\footnotetext{*The first two authors contributed equally to this work.}
\footnotetext{\dag Corresponding author. Email: \nolinkurl{bodong.zhang@utah.edu}}
\footnotetext{\ddag Present address: Department of Pathology, University of California, San Francisco, CA 94143, USA}

\section{Introduction}
Digital histopathology has become integral to cancer diagnosis and research, leveraging hematoxylin and eosin (H\&E) stained tissue sections that are digitized into whole slide images (WSIs) by high-resolution scanners. With increasing computational capacity, deep learning has become a powerful tool for automated disease classification, prognostication, and biomarker development. In practice, because of the gigapixel scale of WSIs, small image patches digitally cropped from WSIs are often used as inputs to deep learning models. Typically, thousands of patches can be extracted from a single slide.  Traditionally, achieving robust performance requires large training sets with fine-grained, patch-level annotations. However, generating such labels is prohibitively time-consuming, costly, and dependent on expert pathology review, creating a major bottleneck for algorithm development at scale. Consequently, strategies that reduce or eliminate the need for dense patch-level supervision have become essential for advancing computational pathology.

% \begin{figure}[!t]
% \centering
% \includegraphics[scale=.4]{WSI_patch_example2.png}
% \caption{Example of a patch cropped from whole slide image (WSI).}
% \label{WSI_patch_example}
% \end{figure}

In recent years, multiple instance learning (MIL) has emerged as an effective strategy to overcome the need for fine-grained annotations in WSI analysis. MIL is a weakly supervised learning paradigm in which labels are provided at the bag level, such as a WSI, rather than for individual instances, i.e., image patches. This allows models to learn discriminative patterns directly from slide-level diagnoses, substantially reducing the burden of patch-level manual labeling while preserving clinically meaningful supervision. %It is particularly valuable when fine-grained labeling is difficult or expensive. 
Under the standard MIL assumption, a bag is labeled positive if at least one instance (patch) within it is positive, and negative if all instances (patches) are negative. It is worth noting that different bags may contain varying numbers of instances. In positive bags, it is often unclear which and how many instances are truly positive. Because individual instance labels are not required, MIL can better tolerate the ambiguity of instance labels.

In histopathology slide classification applications of MIL, WSIs are first divided into small patches (instances), and pretrained encoders are used to extract feature embeddings for each patch. All features of patches from a single WSI are then grouped into a bag, with the slide-level label serving as the bag label. The MIL model takes these patch-level features rather than the raw image patches as input, and is trained to classify each bag as positive or negative (e.g., tumor present or absent).

Most existing MIL approaches directly use patch features as inputs and focus on feature aggregation. However, while the quality of patch features plays a crucial role in determining the MIL performance, it is often overlooked. In many studies that propose new MIL models for histopathology image analysis, ImageNet pretrained encoders are still commonly used as default feature extractors. \cite{shao2021transmil,zhang2022dtfd,yang2024mambamil} The substantial difference between natural images and histopathology images introduces a significant domain shift, which can lead to suboptimal results. Recently, multiple pathology foundation models have been pretrained on large collections of histopathology images and are available for public use. \cite{xu2024whole,chen2024uni,lu2024visual,vorontsov2024foundation,filiot2024phikon,nechaev2024hibou} These foundation models help mitigate domain shift and produce more representative patch embeddings. Nevertheless, since foundation models are trained on external datasets, a degree of domain shift still remains relative to the target domains of local datasets. New studies on finetuning foundation models to improve the performance of the downstream task demonstrate the importance of training on local datasets. \cite{roth2024low,campanella2025real}

In feature representation learning, self-supervised contrastive learning such as SimCLR \cite{chen2020simple} and MoCo \cite{he2020momentum} has emerged as a powerful paradigm for learning high-quality feature representations without relying on human-annotated labels. The central idea is to pull together feature representations derived from different augmented views of the same image while pushing apart representations from different images. %By doing so, the model learns to identify intrinsic semantic similarities rather than relying on explicit supervision. This process encourages the encoder to capture invariant features that remain consistent across a range of transformations, such as changes in brightness, saturation, cropping, or rotation, while still being sensitive to meaningful differences between distinct instances. 
As a result, self-supervised contrastive learning enables the pretrained encoder to generate robust and discriminative representations that can be effectively used in various downstream tasks. \cite{tellez2018gigapixel}

Building upon the principles of self-supervised contrastive learning, supervised contrastive learning (SupCon) extends the contrastive framework to fully leverage patch-level label information. While self-supervised contrastive learning treats each sample as its own class, SupCon introduces supervision by grouping samples that share the same patch-wise class label. Specifically, SupCon aims to minimize the distance between feature representations of patches belonging to the same class (intra-class similarity) and maximize the distance between those from different classes (inter-class dissimilarity). This formulation provides a more structured feature space, improving class separability and representation robustness. Compared to standard supervised learning such as end-to-end training with cross-entropy loss, SupCon encourages a richer embedding geometry and often leads to better generalization and transferability. However, unlike self-supervised approaches that require no annotation, SupCon depends on patch-level labels to define positive pairs, which constrains its applicability in multiple instance learning settings that have limited labeled data.

Inspired by supervised contrastive learning, we propose a novel weakly supervised contrastive learning method called WeakSupCon for feature representation learning in MIL settings. We employ a multi-task learning framework for encoder pretraining. Specifically, image patches are divided into two subsets based on their bag-level labels, and distinct loss terms are assigned to each subset. Unlike self-supervised contrastive learning where positive and negative patch features are mixed due to the lack of label information during representation learning, our method can theoretically separate positive and negative patch features even though the truly positive patches are unknown in MIL scenarios. Through downstream experiments using MIL models on three histopathology datasets, our WeakSupCon model outperforms self-supervised contrastive learning methods and produces more discriminative patch features that lead to improved MIL classification performance. This paper extends our previous conference work \cite{zhang2025weaksupcon} presented in the MICCAI 2025 Workshop on Efficient Medical AI. In this extended version, we present a more comprehensive description of our method and deeper experimental results. Instead of evaluating features generated by different encoder pretraining methods using only a single MIL model, we now assess the learned features across multiple MIL models. Additionally, we include detailed visualizations of feature distributions to evaluate the effectiveness of different feature representation learning strategies. To further investigate the contribution of each loss term, we also conduct additional ablation studies. Experiments across multiple datasets and MIL architectures confirm the robustness and effectiveness of our proposed method.

\section{Related Works}
In this section, we will briefly review the methods in multiple instance learning (MIL), foundation models, self-supervised contrastive learning, and supervised contrastive learning.

\subsection{Multiple Instance Learning}

Multiple instance learning (MIL) has become a widely adopted framework in histopathology image analysis due to its ability to learn from weakly labeled data. Early MIL approaches focus primarily on simple feature aggregation strategies. Max-pooling MIL selects the highest instance score within the bag to represent the bag-level score, whereas mean-pooling MIL computes the bag representation as the average of all instance embeddings. \cite{wang2019comparison} Although both methods are computationally efficient, max-pooling discards information from all other instances, and mean-pooling treats all instances equally, which can be suboptimal.

To address these limitations, attention-based MIL (AB-MIL) \cite{ilse2018attention,tourniaire2021attention} was introduced, where learnable attention weights assign varying importance to different instances within a bag. The bag-level representation becomes a weighted average of instance-level features. This mechanism enables the model to focus on more informative patches rather than treating all instances equally.

The Double-Tier Feature Distillation MIL (DTFD-MIL) \cite{zhang2022dtfd} further enhances training especially in situations where the number of bags is limited, but each positive bag usually contains an adequate number of positive instances. In DTFD-MIL, instances in each bag are randomly split into multiple pseudo-bags, which inherit the original bag label under the assumption that most pseudo-bags generated from positive bags still contain positive instances. In the first tier, pseudo-bags are treated as original bags in MIL training. In the second tier, pseudo-bag-level feature representations are regarded as instance-level representations for the original bags during optimization.

\subsection{Foundation Models}

The development of foundation models has greatly benefited medical image analysis by leveraging large-scale pretraining, where pretrained encoders provide representative features for various downstream tasks, including MIL. This is particularly valuable for small research groups with limited computing resources for large-scale training. One of the most commonly used foundation models is ImageNet-pretrained encoders \cite{deng2009imagenet}, even for histopathology tasks. However, because ImageNet consists of natural images, using such encoders introduces a substantial domain shift for histopathology applications.

To mitigate this gap, domain-specific pathology foundation models have been developed. Early models were trained on relatively small datasets. For example, PathDino \cite{alfasly2023rotation} was trained on approximately 6 million patches extracted from over 11,000 TCGA whole slide images (WSIs) using 4 A100 GPUs. Recent state-of-the-art pathology foundation models were trained on much larger datasets and demonstrate superior performance. Prov-GigaPath \cite{xu2024whole} was pretrained on roughly 1.3 billion pathology patches from more than 170,000 WSIs using 64 A100 GPUs. The dataset covers 31 tissue types, ensuring strong generalization ability. UNI2-h \cite{chen2024uni} is another recent pathology foundation model pretrained on over 200 million patches cropped from more than 350,000 diverse H\&E and IHC WSIs. Although the GPU count for UNI2-h is not reported, its predecessor, UNI, was trained with 32 A100 GPUs.

Despite continuous improvements in histopathology foundation models, domain shift is not fully resolved. When local datasets belong to new domains or differ significantly from the pretraining data, additional feature representation learning on the local dataset remains necessary.

\subsection{Self-Supervised Contrastive Learning}

Self-supervised contrastive learning \cite{manoochehri2024sra} \cite{zhang2022stain} \cite{zhang2023class} provides an effective way to learn feature representations without labels. Momentum Contrast (MoCo) \cite{he2020momentum} proposed a new pretraining method using a dynamic dictionary of negative samples and a momentum-updated encoder. This approach enables efficient training with a large number of negative examples while using small batch sizes. SimCLR \cite{chen2020simple} is another milestone and one of the most popular frameworks. It demonstrates that large batch sizes and strong augmentations improve representation learning without using a memory mechanism. SimCLR also appends a nonlinear projection head on top of the encoder before computing the contrastive loss, which proves beneficial. Those ideas were later incorporated into subsequent MoCo versions. \cite{chen2020improved} \cite{xie2021self}

\begin{figure}[H]
\centering
\includegraphics[scale=.37]{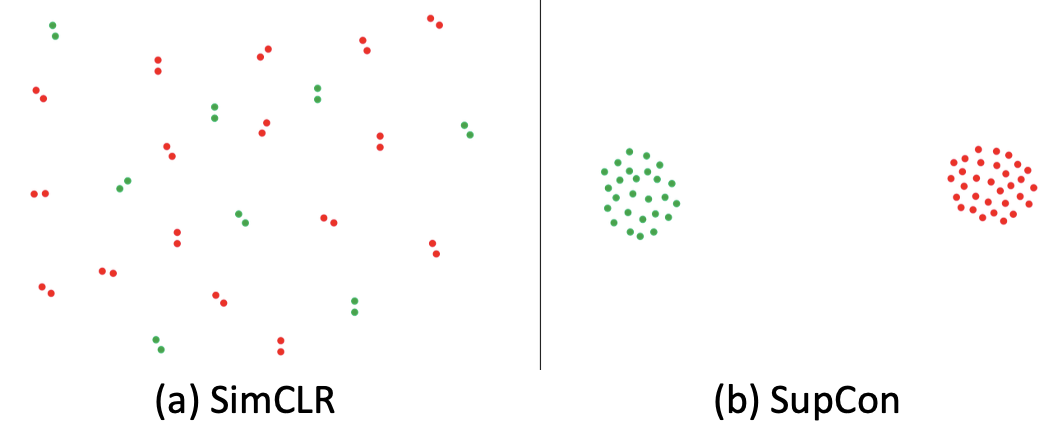}
\caption{Visualizations of theoretical feature distributions after (a) SimCLR and (b) SupCon training. Green dots indicate negative (i.e., no cancer) patches, while red dots indicate positive (i.e., cancer-containing) patches.}
\label{fig_supcon}
\end{figure}

In SimCLR, a batch of N original samples is augmented twice, yielding 2N augmented samples. The contrastive loss between samples i and j (if they form a positive pair) can be formulated as:

\begin{equation}
\ell_{i,j}=-\log{\frac{\exp(sim(z_{i},z_{j})/\tau)}{\sum_{k=1}^{2N}\mathbf{1}_{[k\not=i]}\exp(sim(z_{i},z_{k})/\tau)}}
\label{eq:SimCLR_term}
\end{equation}

where $z_i$ and $z_j$ are the features of samples i and j generated by the encoder followed by the projection head. The $sim(u,v)=u^{\top}v/\|u\|\|v\|$. The $z_a$ and $z_b$ forms a positive pair if they originate from the same sample, and a negative pair if they originate from different original samples. An example of feature distribution learned through SimCLR training is illustrated in Fig.~\ref{fig_supcon}(a).

%\begin{figure}[!t]

\begin{figure*}[!t]
\centering
\includegraphics[width=0.85\textwidth]{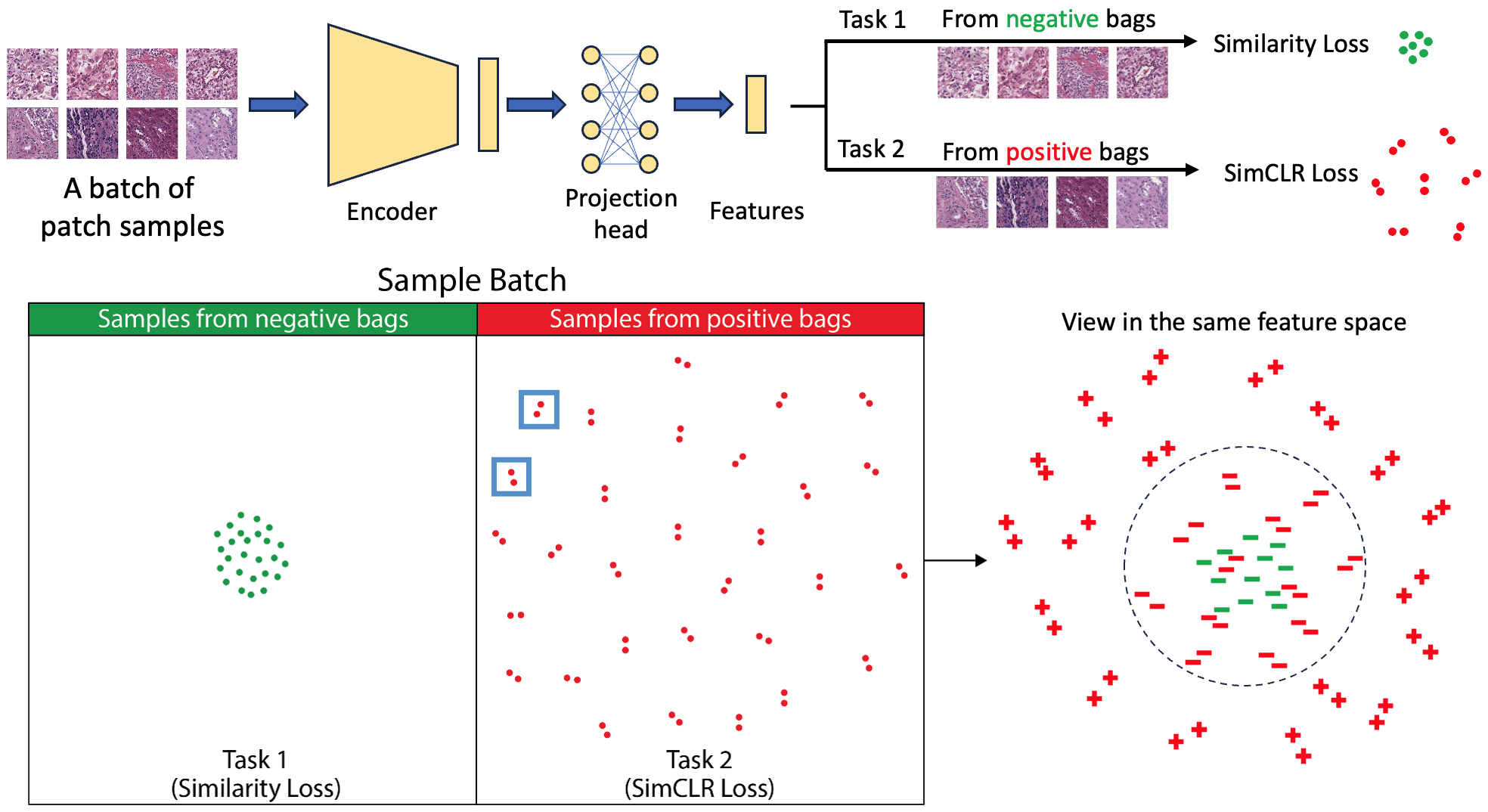}
\caption{Schematic depiction of the main concept of WeakSupCon. The patch-level features generated by the encoder are split into two subsets for separate contrastive learning tasks based on their bag-level labels. Green dots are features from negative bags. Red dots are features from positive bags. In task 1, our proposed Similarity Loss is applied to patches from negative bags, which form a tight cluster. In task 2, the SimCLR Loss tries to maximize the distance between features from different original patches. Dots in the same blue box are from different augmentations of the same original patch. The bottom right part reveals an anticipated feature distribution of positive ('+') and negative ('-') patches (instances) when viewing in the same feature space. }\label{fig_weaksupcon}
\end{figure*}

\subsection{Supervised Contrastive Learning}

Building on the success of self-supervised contrastive learning, supervised contrastive learning (SupCon) extends SimCLR by leveraging instance-level label information. In SupCon, samples with the same class label are treated as positive pairs, while those from different classes are treated as negative pairs. The SupCon loss is defined as:

\begin{equation}
\mathcal{L}^{sup} = \sum_{i \in I} \frac{-1}{|P(i)|} \sum_{p \in P(i)}{\log \frac{\exp(z_{i}\cdot z_{p}/\tau)}{\sum_{a \in A(i)}{\exp{(z_{i} \cdot z_{a} / \tau)}}}}
\label{eq:SuoConLoss}
\end{equation}

where P(i) represents the set of indices sharing the same label as sample i, and A(i) denotes the set of all indices except i. A pair $(z_i, z_p)$ is a positive pair if the two samples share the same label. $\tau$ is a temperature hyperparameter. The theoretical feature distribution learned using SupCon is shown in Fig.~\ref{fig_supcon}(b). The SupCon loss effectively minimizes intra-class variance while maximizing inter-class separability in the feature space, enabling more discriminative representations than self-supervised contrastive learning.

In experiments on datasets with instance-level labels, including ImageNet \cite{deng2009imagenet}, CIFAR-10, and CIFAR-100 \cite{krizhevsky2009learning}, the features generated by SupCon were frozen, and a linear classifier was trained using the learned features as input. The results show that SupCon outperforms self-supervised contrastive learning and even end-to-end training by cross-entropy loss. However, the requirement of instance-level labels limits its direct application in MIL.

\section{Method}
\subsection{An overview of WeakSupCon}
Despite the challenge of applying SupCon in MIL, we draw two key insights. First, incorporating label information during feature representation learning has the potential to improve feature quality. Second, even though end-to-end MIL training from images to final classification is an interesting topic \cite{campanella2019clinical, wang2023iteratively}, it is extremely challenging, especially when a single bag contains thousands of patches in histopathology applications. However, SupCon demonstrates that decoupling representation learning from downstream classification can still achieve strong performance. Motivated by these observations, we propose a new weakly supervised contrastive learning method, termed WeakSupCon, to learn feature embeddings by utilizing bag-level labels in MIL settings.

Fig.~\ref{fig_weaksupcon} illustrates the main idea of Weakly Supervised Contrastive Learning (WeakSupCon). First, a batch of image patches is randomly sampled from the entire dataset, including patches from both negative and positive bags (slides). The patches are passed through an encoder to obtain latent feature representations. Following the design of self-supervised contrastive learning frameworks (such as SimCLR \cite{chen2020simple}) and supervised contrastive learning (SupCon \cite{khosla2020supervised}), the resulting features are further transformed through a projection head before loss computation. We then separate the projected features into two subsets according to their bag-level labels. A multi-task learning strategy is applied by assigning different loss optimization tasks to the two subsets. The first subset consists of patch features derived from negative bags, while the second contains patch features originating from positive bags. 
%To do: multi task learning. First, second

In task 1, we process patch features from negative bags. By definition, a bag receives a negative label only when all its instances (patches) are negative. Inspired by SupCon, which forces features with the same label to cluster together, we propose a Similarity Loss that encourages features of negative patches to minimize their pairwise distances, thereby increasing their similarity. However, unlike SupCon, where instance-level labels are available, in multiple instance learning (MIL) settings, only a portion of instances (patches) in positive bags (slides) are positive. It is even likely that most instances in positive bags are actually negative, making it unclear which patches in positive bags are truly positive. As a result, the Similarity Loss does not include features from positive bags, nor does it attempt to maximize the distance between features of negative and positive patches. This design explicitly avoids introducing incorrect supervisory signals that would arise from assuming positive patches can be reliably identified in a weakly supervised setting.

In task 2, we process patch features from positive bags. In these bags, a portion of the patches is positive. However, the number of positive patches varies across individual bags/slides. Because the fraction of positive patches is unknown and patch-level labels are not available, applying supervised contrastive learning is impractical in this context. Instead, we apply the self-supervised SimCLR Loss to features from positive bags. Because the SimCLR Loss aims to maximize the distance between features derived from different original patches (whether positive or negative), it promotes greater diversity in the feature space and ensures that potential positive patches are not inadvertently collapsed with negative patches. 

During training, both tasks share the same encoder and projection head. Although the patches are divided into two subsets for separate tasks, the encoder is jointly optimized for both tasks. The total loss is the sum of the Similarity Loss and the SimCLR Loss. As shown in Fig.~\ref{fig_weaksupcon}, the feature distributions of the two subsets can be viewed within a shared feature space. In the training set, patch features from negative bags cluster together due to the Similarity Loss, while features from positive bags display greater variability because of the SimCLR Loss. If the encoder generalizes well, even though the true negative patches within positive bags are not directly used in the Similarity Loss, their features should still resemble those from negative bags, as they share the same label and similar characteristics. For truly positive patches in positive bags, this effect does not occur, and their features tend to diverge from negative ones because the SimCLR Loss increases the distance between features of different patches. Consequently, as illustrated in Fig.~\ref{fig_weaksupcon}, even without patch-level labels in MIL, the loss design inherently promotes separation between positive and negative patches. Furthermore, positive patches exhibit greater diversity in the feature space, making them more likely to receive higher attention weights in downstream MIL processes.

\subsection{Loss Functions}
Our WeakSupCon loss consists of two components: the Similarity Loss and the SimCLR Loss. The Similarity Loss is created by modifications on the SupCon loss. The SupCon loss can be rewritten as follows:

% \begin{equation}
% \begin{aligned}
% \mathcal{L}^{sup} &= \sum_{i \in I} \frac{-1}{|P(i)|} \sum_{p \in P(i)}{\log \frac{\exp(z_{i}\cdot z_{p}/\tau)}{\sum_{a \in A(i)}{\exp{(z_{i} \cdot z_{a} / \tau)}}}} \\
%   &= \sum_{i \in I} \frac{-1}{|P(i)|} \sum_{p \in P(i)}{\log (\exp(z_{i}\cdot z_{p}/\tau)) - \log {\sum_{a \in A(i)}{\exp{(z_{i} \cdot z_{a} / \tau)}}}      } \\
%   &= \sum_{i \in I} \frac{-1}{|P(i)|} \sum_{p \in P(i)}{(z_{i}\cdot z_{p}/\tau - \log {\sum_{a \in A(i)}{\exp{(z_{i} \cdot z_{a} / \tau)}}})} \\
% \label{eq:SupConLoss_deduction}
% \end{aligned}
% \end{equation}

\begin{equation}
%\begin{aligned}
\begin{alignedat}{1}
&\mathcal{L}^{sup} = \sum_{i \in I} \frac{-1}{|P(i)|} \sum_{p \in P(i)}{\log \frac{\exp(z_{i}\cdot z_{p}/\tau)}{\sum_{a \in A(i)}{\exp{(z_{i} \cdot z_{a} / \tau)}}}} \\
&= \sum_{i \in I} \frac{-1}{|P(i)|} \sum_{p \in P(i)}{\log (\exp(\frac{z_{i}\cdot z_{p}}{\tau})) - \log {\sum_{a \in A(i)}{\exp{(\frac{z_{i} \cdot z_{a}}{\tau})}}}      } \\
&= \sum_{i \in I} \frac{-1}{|P(i)|} \sum_{p \in P(i)}{(z_{i}\cdot z_{p}/\tau - \log {\sum_{a \in A(i)}{\exp{(z_{i} \cdot z_{a} / \tau)}}})} \\
\end{alignedat}
\label{eq:SupConLoss_deduction}
%\end{aligned}
\end{equation}

% \begin{equation}
% \begin{alignedat}{1}
% \mathcal{L}^{sup} = \sum_{i \in I} \frac{-1}{|P(i)|} \sum_{p \in P(i)} \cdots \\
% = \sum_{i \in I} \frac{-1}{|P(i)|} \sum_{p \in P(i)} \cdots \\
% = \sum_{i \in I} \frac{-1}{|P(i)|} \sum_{p \in P(i)} \cdots
% \end{alignedat}
% \label{eq:SupConLoss_deduction}
% \end{equation}

The first component, $z_{i}\cdot z_{p}/\tau$, measures the similarity between features that share the same label. By minimizing the SupCon loss, features with identical labels become more similar in the feature space. The second component includes computing the similarity between different labels. Minimizing the total loss increases the separation between features with distinct labels because of the negative sign in the second component.

In our WeakSupCon, since all features in the first group share the same label, we directly adopt only the first component of the SupCon loss to cluster negative features together:

\begin{equation}
\mathcal{L}^{Similarity} = \sum_{i \in Neg} \frac{-1}{|Neg|} \sum_{j \in Neg, j \neq i}{z_{i}\cdot z_{j}/\tau}
\label{eq:SimilarityLoss}
\end{equation}
where Neg represents the set of samples in negative bags. 

The SimCLR Loss remains the same as defined in \cite{chen2020simple}:

\begin{equation}
\mathcal{L}^{SimCLR} = \sum_{i \in Pos} \ell_{i, p(i)}
\label{eq:SimCLRLoss}
\end{equation}
where Pos represents the set of samples in positive bags, and p(i) represents a feature of another augmented image that shares the same original image as i. The $\ell_{i, p(i)}$ is defined in \eqref{eq:SimCLR_term}. 

The total loss is a weighted sum of the Similarity Loss and the SimCLR Loss:

\begin{equation}
\mathcal{L}^{WeakSupCon} = \alpha\mathcal{L}^{Similarity} + \mathcal{L}^{SimCLR}
\label{eq:WeakSupConLoss}
\end{equation}

where $\alpha$ represents the weight assigned to the Similarity Loss. When $\alpha=1$, all samples are given equal weights in the computation, regardless of whether they belong to negative or positive bags.

\subsection{Downstream Multiple Instance Learning (MIL)}
After pretraining, we use the encoder to extract features from all image patches. These features are then frozen and used as inputs for downstream MIL tasks. %The quality of the extracted features is crucial for the performance of the MIL model.

Our WeakSupCon framework provides a theoretically clear separation between negative and positive patches while encouraging greater variations among positive patches in the feature space. This variance facilitates the assignment of higher and more informative attention weights during MIL training.

% \caption{Evaluation of pretraining methods. !!!! Multiple instance learning results comparison by using different encoders, including self-supervised contrastive learning models (MoCo v3, SimCLR), supervised contrastive learning model (SupCon), and our WeakSupCon. Both DTFD-MIL and AB-MIL were tested independently as downstream MIL models for evaluation on the Camelyon16 dataset, renal vein thrombosis (RVT) dataset, and kidney metastasis dataset.}\label{tab1_contrastive_learning}
\begin{table*}[t]
\centering
\caption{Evaluation of pretraining methods, including self-supervised contrastive learning models (MoCo v3, SimCLR), supervised contrastive learning model (SupCon), and our WeakSupCon. DTFD-MIL and AB-MIL were experimented independently as downstream MIL models for evaluation of learned features on the Camelyon16 dataset, renal vein thrombosis (RVT) dataset, and kidney metastasis dataset.}\label{tab1_contrastive_learning}
\setlength{\tabcolsep}{3pt}
\begin{tabular}{|l|l|l|l|l|l|l|}
\hline
 & \multicolumn{3}{|c|}{DTFD-MIL Results on Camelyon16 dataset} & \multicolumn{3}{|c|}{AB-MIL Results on Camelyon16 dataset} \\ 
\hline
\textbf{Encoder} &  \textbf{Balanced acc} & \textbf{Accuracy} & \textbf{AUC} &  \textbf{Balanced acc} & \textbf{Accuracy} & \textbf{AUC}\\
\hline
MoCo v3  &  0.9051$\pm$0.0200 & 0.9199$\pm$0.0195 & 0.9238$\pm$0.0050 &  0.8939$\pm$0.0291 & 0.9044$\pm$0.0249 & 0.9061$\pm$0.0264\\
\hline
SimCLR  &  0.8928$\pm$0.0050  & 0.9095$\pm$0.0089 & 0.9130$\pm$0.0048 &  0.8567$\pm$0.0099  & 0.8811$\pm$0.0044 & 0.8858$\pm$0.0129\\
\hline
SupCon  &  0.8760$\pm$0.0191  & 0.9018$\pm$0.0161 & 0.8792$\pm$0.0077 &  0.8705$\pm$0.0042  & 0.8966$\pm$0.0044 & 0.8519$\pm$0.0024\\
\hline
WeakSupCon  &  {\bfseries 0.9265}$\pm$0.0036  & {\bfseries 0.9431}$\pm$0.0044 & {\bfseries 0.9694}$\pm$0.0018 &  {\bfseries 0.9318}$\pm$0.0042  & {\bfseries 0.9431}$\pm$0.0044 & {\bfseries 0.9699}$\pm$0.0009\\
\hline
 & \multicolumn{3}{|c|}{DTFD-MIL Results on RVT dataset} & \multicolumn{3}{|c|}{AB-MIL Results on RVT dataset} \\ 
\hline
\textbf{Encoder} &  \textbf{Balanced acc} & \textbf{Accuracy} & \textbf{AUC} &  \textbf{Balanced acc} & \textbf{Accuracy} & \textbf{AUC}\\
\hline
MoCo v3  &  0.7012$\pm$0.0466  & 0.7241$\pm$0.0345 & 0.7610$\pm$0.0556  &  0.7256$\pm$0.0325  & 0.7471$\pm$0.0199 & 0.7609$\pm$0.0372\\
\hline
SimCLR  &  0.6785$\pm$0.0239  & 0.7012$\pm$0.0526 & 0.7592$\pm$0.0466 &  0.7096$\pm$0.0522  & 0.7126$\pm$0.0526 & 0.7340$\pm$0.0540\\
\hline
SupCon  &  0.7281$\pm$0.0368  & 0.7356$\pm$0.0527 & 0.8266$\pm$0.0304 &  0.7534$\pm$0.0325  & 0.7816$\pm$0.0199 & 0.8333$\pm$0.0051\\
\hline
WeakSupCon  &  {\bfseries 0.8014}$\pm$0.0130  & {\bfseries 0.8046}$\pm$0.0199 & {\bfseries 0.8771}$\pm$0.0177 &  {\bfseries 0.7980}$\pm$0.0307  & {\bfseries 0.7931}$\pm$0.0000 & {\bfseries 0.8788}$\pm$0.0315\\
\hline
 & \multicolumn{3}{|c|}{DTFD-MIL Results on kidney metastasis dataset} & \multicolumn{3}{|c|}{AB-MIL Results on kidney metastasis dataset} \\ 
\hline
\textbf{Encoder} &  \textbf{Balanced acc} & \textbf{Accuracy} & \textbf{AUC} &  \textbf{Balanced acc} & \textbf{Accuracy} & \textbf{AUC}\\
\hline
MoCo v3  &  0.8633$\pm$0.0135  & 0.8261$\pm$0.0000 & 0.8982$\pm$0.0048 &  0.8605$\pm$0.0044  & 0.8333$\pm$0.0332 & 0.8803$\pm$0.0141\\
\hline
SimCLR  &  0.8939$\pm$0.0000  & 0.8478$\pm$0.0000 & 0.9192$\pm$0.0088 &  0.8889$\pm$0.0087  & 0.8406$\pm$0.0125 & 0.9192$\pm$0.0110\\
\hline
SupCon  &  0.8858$\pm$0.0000  & {\bfseries 0.8696}$\pm$0.0000 & 0.9176$\pm$0.0016 &  0.8807$\pm$0.0088  & {\bfseries 0.8623}$\pm$0.0126 & 0.9223$\pm$0.0027\\
\hline
WeakSupCon  &  {\bfseries 0.9091}$\pm$0.0000  & {\bfseries 0.8696}$\pm$0.0000 & {\bfseries 0.9277}$\pm$0.0016 &  {\bfseries 0.9040}$\pm$0.0088  & {\bfseries 0.8623}$\pm$0.0126 & {\bfseries 0.9246}$\pm$0.0013\\
\hline

\end{tabular}
\end{table*}

\section{Experiments}
\subsection{Datasets}
We conducted experiments on three datasets: (1) Camelyon16 dataset, (2) renal vein thrombosis (RVT) dataset, which includes slides from nephrectomy cases of clear cell renal cell carcinoma (ccRCC) who were diagnosed with RVT from our institution, and (3) kidney metastasis dataset, with slides from nephrectomy cases with ccRCC who developed kidney cancer metastasis from our institution. All three datasets consist of histopathology whole slide images (WSIs) stained with hematoxylin and eosin (H\&E). The slides in each dataset were divided into training, validation, and test sets.

Camelyon16 \cite{bejnordi2017diagnostic} is a publicly available multi-institutional dataset that contains 400 whole slide histopathology images. It includes slides of breast cancer lymph node metastases (positive bags) and normal lymph nodes without metastatic breast cancer (negative bags). The amount of cancer in the positive WSIs is highly variable and in some cases only comprises a small area. In these slides, there are only a small portion of positive cancer patches, while the remaining patches are negative patches which consist of lymphocytes and other benign cells. Slide-level labels are provided with the dataset. The slides were split into training set and test set. In the training set, there are 270 slides in total. We randomly parsed 10\% of them into a validation set. Using the implementation code from \cite{zhang2022dtfd} to crop patches from all slides, a total of 2,292,709 patches were extracted from the final training set.

The renal vein thrombosis (RVT) dataset \cite{sirohi2022histologic} contains case-level (patient-level) labels. Each case typically consists of multiple slides. A case is labeled positive (with RVT) based on the label in the pathology report. %Multiple sildes represent the ccRCC inside the kidney. 
The dataset was separated into 105, 20, and 29 cases to generate training, validation, and test sets. In the training set, there are 31 cases with RVT, from which 695,439 patches were collected, and 74 cases without RVT, from which 1,518,872 patches were cropped.

The kidney metastasis dataset \cite{sirohi2022histologic} also contains case-level labels and all cases received a diagnosis of ccRCC. The metastases occurred to the regional lymph nodes, which were removed during surgery. In total, there are 57, 36, and 46 cases in the training, validation, and test sets. All WSIs from the metastatic cases are from the primary tumor in the kidney. The training set includes 42 metastatic cases, from which 930,939 patches were cropped from 393 slides. There are also 15 non-metastatic cases, from which 383,725 patches were extracted from 107 slides of primary ccRCC.

\subsection{Experimental Settings}
For feature representation learning, we pretrained encoders on all training patches within each dataset. Here, we applied our proposed WeakSupCon method, as well as self-supervised contrastive learning methods including SimCLR and MoCo v3, and supervised contrastive learning (SupCon) for comparison.

To enable SupCon, which requires patch-level labels, we assigned pseudo-labels to all patches: patches from positive bags were treated as positive, while patches from negative bags were assigned negative labels. Due to our computational resource constraints, we used a batch size of 512 and ResNet-18 as the backbone for all contrastive learning methods. Compared with ViT-tiny, we found that ResNet-18 was easier to train and consistently achieved more than 10\% higher accuracy. Each experiment typically requires around 35 GB of GPU memory on a single NVIDIA RTX A6000 GPU.

After pretraining on the training set, we extracted patch features in training, validation, and test sets using the trained encoders. The features were frozen and used as inputs for downstream MIL tasks. We independently evaluated the feature representations using DTFD-MIL and AB-MIL. Performance on the validation set was monitored during each epoch. The epoch achieving the highest AUC on the validation set was selected for final evaluation on the test set. The performance of the models was assessed using balanced accuracy, accuracy, and AUC on the classification endpoint by running three times.

% \begin{figure*}[!t]
% \centering
% \includegraphics[width=0.7\textwidth]{PCA_vis_training_set.png}
% \caption{PCA visualizations of features on training set, red dots denote features from positive bags, green dots denote features from negative bags. The horizontal and vertical axes represent the first two dimensions in PCA. }\label{fig_PCA_training_set}
% \end{figure*}

% \begin{figure*}[!t]
% \centering
% \includegraphics[width=0.7\textwidth]{PCA_vis_test_set.png}
% \caption{PCA visualizations of features on test set, red dots denote features from positive bags, green dots denote features from negative bags. The horizontal and vertical axes represent the first two dimensions in PCA.}\label{fig_PCA_test_set}
% \end{figure*}

\begin{figure*}[!t]
\centering
\includegraphics[width=0.7\textwidth]{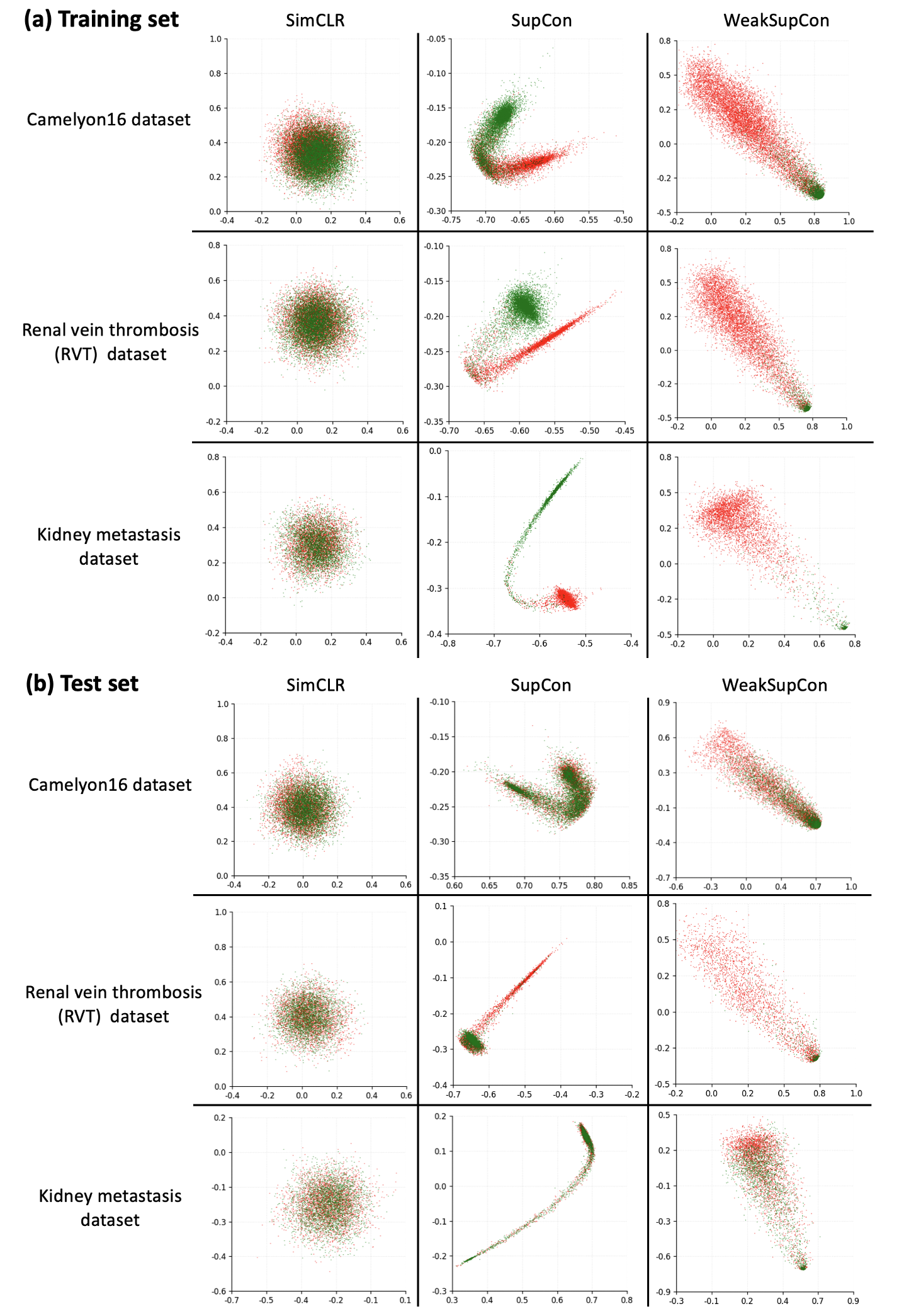}
\caption{PCA visualizations of features on (a) training set and (b) test set, red dots denote features from positive bags, green dots denote features from negative bags. The horizontal and vertical axes represent the first two dimensions in PCA. The features were downsampled for better visualization.}\label{fig_PCA_training_test_set}
\end{figure*}

\subsection{Model performance}
Table~\ref{tab1_contrastive_learning} presents the comparison between our WeakSupCon and other contrastive learning models, including MoCo v3, SimCLR, and SupCon, using DTFD-MIL and AB-MIL as downstream MIL models. Overall, the features generated by our WeakSupCon consistently achieve the best performance in both DTFD-MIL and AB-MIL across the three datasets. It is worth mentioning that both SupCon and our WeakSupCon are based on SimCLR, with modifications to the loss design. Therefore, the comparison among these three methods is fair. We observed approximately 1.5–7\% improvements with DTFD-MIL and 1.5–4.5\% improvements with AB-MIL in balanced accuracy over other contrastive learning methods.

An interesting observation is that on the renal vein thrombosis (RVT) dataset, SupCon demonstrates better performance than MoCo v3 and SimCLR in all metrics. In contrast, we observed that SupCon performs worse than the self-supervised methods SimCLR and MoCo v3 on the Camelyon16 dataset using DTFD-MIL. According to \cite{shao2021transmil} and \cite{zhang2022dtfd}, it is estimated that the positive regions in positive slides of the Camelyon16 dataset occupy only a small portion (less than 10\%) of the tissue area. Consequently, in SupCon, most pseudo patch labels for patches in positive slides are incorrect in the Camelyon16 dataset, which is likely the reason for the low performance.

% In contrast, for the renal vein thrombosis (RVT) dataset, where positive regions occupy a much higher proportion of each slide, the pseudo-labels for patches in positive slides are more accurate. Consequently, SupCon demonstrates better performance than MoCo v3 and SimCLR on this dataset. %Similarly, in the kidney metastasis dataset, SupCon consistently performs better than MoCo v3 with both DTFD-MIL and AB-MIL.

Nevertheless, our WeakSupCon consistently achieves the best performance. The model effectively avoids the challenge of incorrect pseudo-labels while still being able to separate image patches with different labels in the feature space. Moreover, WeakSupCon exhibits relatively small standard deviations across all three metrics (balanced accuracy, accuracy, and AUC), confirming the robustness of our model.

\subsection{Feature Distribution Analysis}
To gain a better understanding of different contrastive learning methods, we visualized the feature distributions of SimCLR, SupCon, and WeakSupCon on both the training and test sets across the three datasets. The Principal Component Analysis (PCA) visualizations of features after the projection head are shown in Fig.~\ref{fig_PCA_training_test_set}. %Red dots denote features of patches from positive bags, while green dots denote features of patches from negative bags. The horizontal axis and vertical axis represent the first two dimensions in PCA. 

For the self-supervised contrastive learning method (SimCLR), label information is not used during training of the encoder. Therefore, the distributions of patch-wise features from positive and negative bags do not exhibit clear separation and overlap substantially after being reduced to two dimensions by PCA. In contrast, for the supervised contrastive learning method (SupCon), we observed a clear difference in the feature distributions of patches from bags with different labels in the training set. %In comparison to the Camelyon16 dataset, it is easier to appreciate the separation of green and red dots in the training sets of the other two datasets, due to their higher proportion of positive instances in positive bags. 
However, the incorrect pseudo-labels assigned to patches in positive bags tend to cause overfitting during training. As illustrated by the feature distributions on the test sets of all three datasets, SupCon often fails to maintain distinct feature distributions between patches from positive and negative bags. This is especially evident in the Camelyon16 and kidney metastasis datasets. The RVT dataset is an exception (shown in Figure 3b, panel RVT/SupCon). A potential explanation is a high proportion of positive patches in cases with RVT, resulting in more accurate pseudo-labels. This observation is consistent with the MIL results on the RVT dataset in Table~\ref{tab1_contrastive_learning}. However, when the proportion of positive instances in positive bags is low, SupCon performs worse than self-supervised methods that do not rely on labels.

In contrast, our WeakSupCon produced clearly distinguishable feature distributions between patches from positive versus negative bags in both the training set and the test set. As expected, in the training set, features from negative bags exhibit strong similarity and cluster together, while those from positive bags vary a lot, as our method separates the embeddings from positive patches to have more diverse feature representations. Furthermore, WeakSupCon demonstrates a wider range among patches along the x-axis (the first principal component) after PCA dimension reduction. Typically, SimCLR exhibits a range of about 0.6 (like from –0.2 to 0.4), SupCon shows a range of approximately 0.3, while WeakSupCon achieves a range of around 1.0. In the test set, some patch features from positive bags appear similar to those from negative bags. However, this is expected, as a portion of patches within positive bags are indeed negative. %Negative patches in Cameleon will consist of benign immune cells while negative patches in ccRCC cases will be patches that are not associated with renal vein thrombosis or metastasis. 
In theory, if a patch from a positive bag is closer to the cluster of green dots, the patch is more likely to be negative. In the Camelyon16 dataset, where positive patches occupy less than 10\% of the positive slides \cite{shao2021transmil, Li2020-da-dsmil}, greater overlap between red and green dots is expected in the test set visualization. In the RVT test set, a larger portion of red dots falls outside the range of green dots, indicating a higher percentage of positive patches in positive cases.

\begin{figure*}[!t]
\centering
\includegraphics[width=0.85\textwidth]{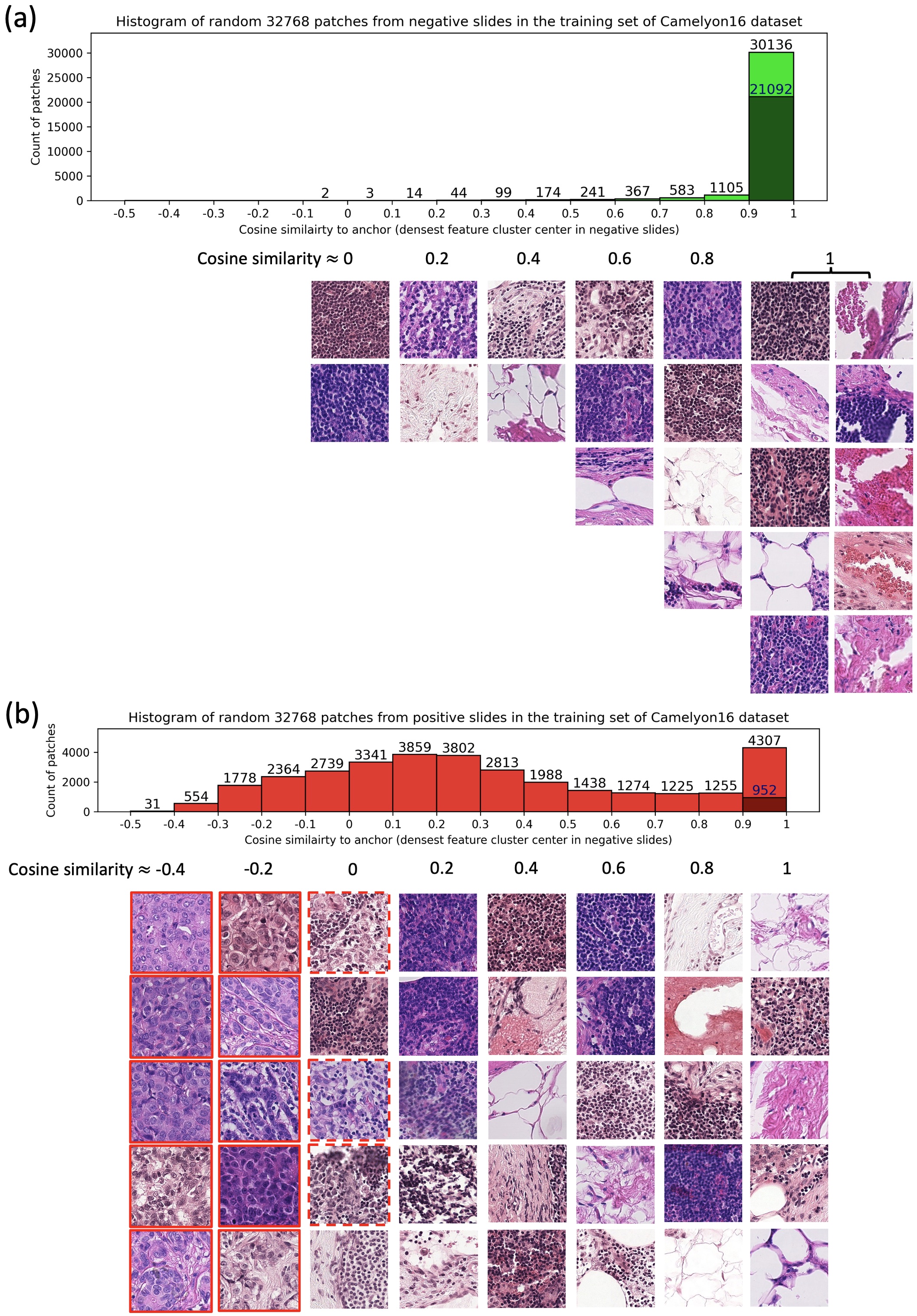}
\caption{Histograms of cosine similarities between anchor (densest feature cluster center in negative slides) and randomly sampled patch features from (a) negative slides and (b) positive slides in the TRAINING set of Camelyon16 dataset generated by WeakSupCon encoder. The dark color in histograms shows the portion with cosine similarity greater than 0.999. The patch examples with red boundaries are positive patches diagnosed by our pathologists. }\label{histogram_train}
\end{figure*}

\begin{figure*}[!t]
\centering
\includegraphics[width=0.80\textwidth]{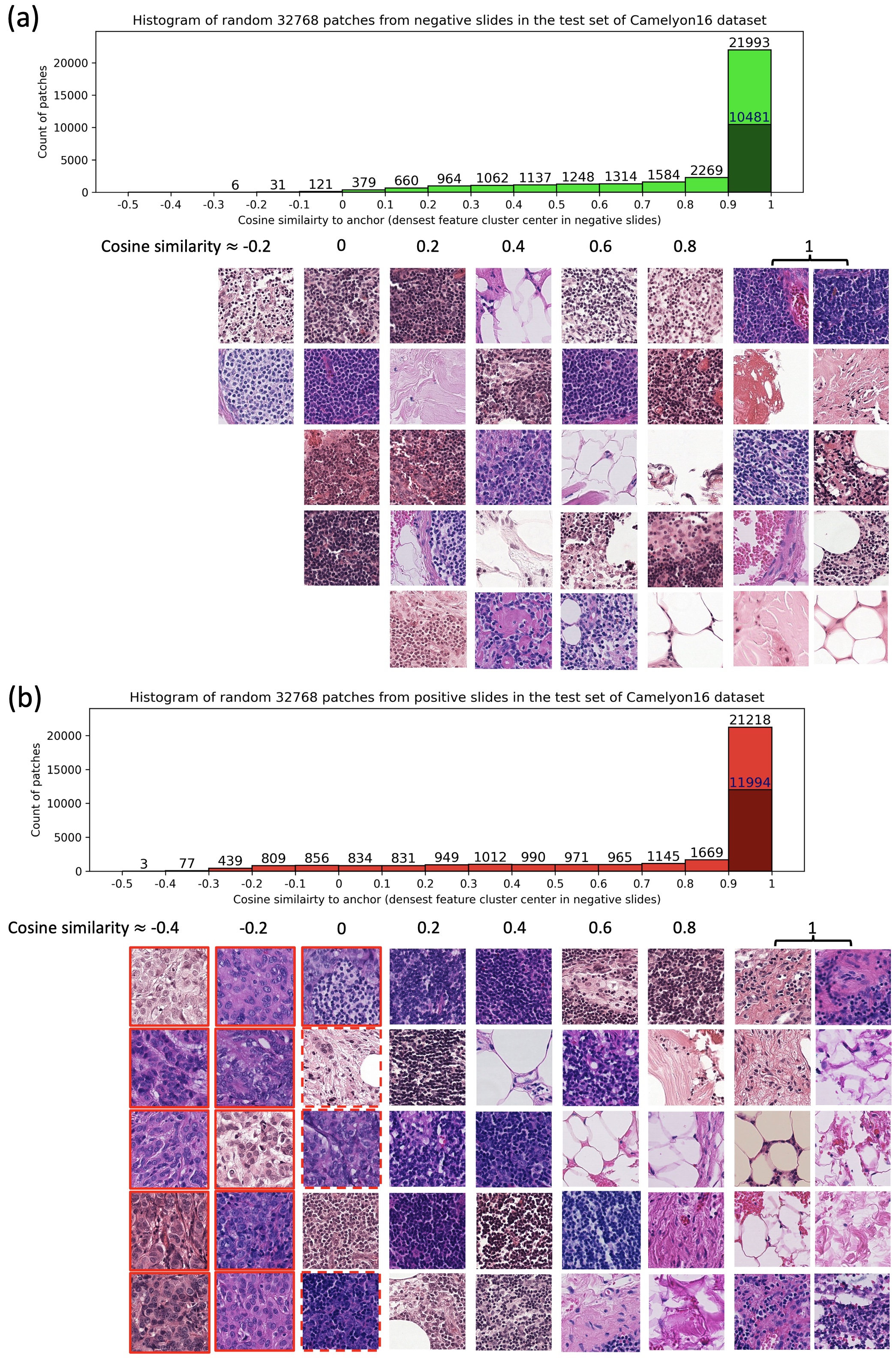}
\caption{Histograms of cosine similarities between anchor (densest feature cluster center in negative slides) and randomly sampled patch features from (a) negative slides and (b) positive slides in the TEST set of Camelyon16 dataset generated by WeakSupCon encoder. The dark color in histograms shows the portion with cosine similarity greater than 0.999. The patch examples with red boundaries are positive patches diagnosed by our pathologists. }\label{histogram_test}
\end{figure*}

\begin{table*}[t]%ht
\centering
\caption{Comparison between our WeakSupCon model and state-of-the-art histopathology foundation models, including Prov-GigaPath and UNI2-h. ImageNet represents ImageNet pretrained encoder.
}\label{tab2_foundation_models}
\setlength{\tabcolsep}{3pt}%narrow the two sides' space in each grid
\begin{tabular}{|l|l|l|l|l|l|l|}
\hline
& \multicolumn{3}{|c|}{DTFD-MIL Results on Camelyon16 dataset} & \multicolumn{3}{|c|}{AB-MIL Results on Camelyon16 dataset} \\ 
\hline
\textbf{Encoder} &  \textbf{Balanced acc} & \textbf{Accuracy} & \textbf{AUC} &  \textbf{Balanced acc} & \textbf{Accuracy} & \textbf{AUC}\\
\hline
ImageNet & 0.8034$\pm$0.0211 & 0.8346$\pm$0.0119 & 0.8530$\pm$0.0186 & 0.7515$\pm$0.0159 & 0.8062$\pm$0.0135 & 0.7599$\pm$0.0110 \\
\hline
Prov-GigaPath  &  {\bfseries 0.9469}$\pm$0.0036  & {\bfseries 0.9586}$\pm$0.0044 & 0.9757$\pm$0.0009 &  {\bfseries 0.9577}$\pm$0.0013  & {\bfseries 0.9638}$\pm$0.0045 & 0.9733$\pm$0.0038\\
\hline
UNI2-h  &  0.9461$\pm$0.0032  & 0.9561$\pm$0.0044 & {\bfseries 0.9782}$\pm$0.0016 &  0.9556$\pm$0.0023  & 0.9612$\pm$0.0000 & {\bfseries 0.9936}$\pm$0.0004\\
\hline
WeakSupCon  &  0.9265$\pm$0.0036  & 0.9431$\pm$0.0044 & 0.9694$\pm$0.0018 &  0.9318$\pm$0.0042  & 0.9431$\pm$0.0044 & 0.9699$\pm$0.0009\\
\hline
& \multicolumn{3}{|c|}{DTFD-MIL Results on RVT dataset} & \multicolumn{3}{|c|}{AB-MIL Results on RVT dataset} \\ 
\hline
\textbf{Encoder} &  \textbf{Balanced acc} & \textbf{Accuracy} & \textbf{AUC} &  \textbf{Balanced acc} & \textbf{Accuracy} & \textbf{AUC}\\
\hline
ImageNet &  0.6330$\pm$0.0238 & 0.6322$\pm$0.0199 & 0.6280$\pm$0.0480 &  0.7037$\pm$0.0304 & 0.7126$\pm$0.0199 & 0.7374$\pm$0.0152 \\
\hline
Prov-GigaPath &  0.7096$\pm$0.0140 & 0.7126$\pm$0.0199 & 0.7576$\pm$0.0101 &  0.7147$\pm$0.1119 & 0.6896$\pm$0.1195 & 0.7273$\pm$0.1488 \\
\hline
UNI2-h &  0.6524$\pm$0.0229 & 0.5977$\pm$0.0199 & 0.6229$\pm$0.0191 &  0.6490$\pm$0.0278 & 0.5862$\pm$0.0345 & 0.6869$\pm$0.0364 \\
\hline
WeakSupCon  &  {\bfseries 0.8014}$\pm$0.0130  & {\bfseries 0.8046}$\pm$0.0199 & {\bfseries 0.8771}$\pm$0.0177 &  {\bfseries 0.7980}$\pm$0.0307  & {\bfseries 0.7931}$\pm$0.0000 & {\bfseries 0.8788}$\pm$0.0315\\
\hline
& \multicolumn{3}{|c|}{DTFD-MIL Results on kidney metastasis dataset} & \multicolumn{3}{|c|}{AB-MIL Results on kidney metastasis dataset} \\ 
\hline
\textbf{Encoder} &  \textbf{Balanced acc} & \textbf{Accuracy} & \textbf{AUC} &  \textbf{Balanced acc} & \textbf{Accuracy} & \textbf{AUC}\\
\hline
ImageNet  &  0.7848$\pm$0.0291  & 0.7246$\pm$0.0251 & 0.8415$\pm$0.0142 &  0.7821$\pm$0.0051  & 0.7319$\pm$0.0450 & 0.8407$\pm$0.0054\\
\hline
Prov-GigaPath  &  0.8862$\pm$0.0076  & 0.8478$\pm$0.0218 & 0.9192$\pm$0.0178 &  0.8834$\pm$0.0118  & 0.8551$\pm$0.0126 & 0.9184$\pm$0.0093\\
\hline
UNI2-h  &  0.8889$\pm$0.0175  & 0.8406$\pm$0.0251 & 0.9161$\pm$0.0185 &  0.8939$\pm$0.0000  & 0.8478$\pm$0.0000 & 0.9215$\pm$0.0036\\
\hline
WeakSupCon  &  {\bfseries 0.9091}$\pm$0.0000  & {\bfseries 0.8696}$\pm$0.0000 & {\bfseries 0.9277}$\pm$0.0016 &  {\bfseries 0.9040}$\pm$0.0088  & {\bfseries 0.8623}$\pm$0.0126 & {\bfseries 0.9246}$\pm$0.0013\\
\hline

\end{tabular}
\end{table*}

\subsection{Quantitative analysis on patch features from WeakSupCon encoder}
To further quantitatively analyze the distribution of features and the advantages introduced by encoder pretraining with WeakSupCon, we used the Camelyon16 dataset as an example.

First, we randomly sampled 32,768 patches from negative slides and positive slides separately in both the training set and the test set. These patches were then transformed into feature representations using the encoder pretrained with WeakSupCon. In theory, features extracted from negative slides tend to cluster together. Therefore, we first sought a representative feature located at the densest region of the cluster formed by features from negative slides. For each feature in negative slides, we counted the number of neighboring features with a cosine similarity greater than 0.999. The feature with the largest number of such neighbors was selected as an anchor and referred to as the densest feature cluster center in negative slides. We then calculated the cosine similarity between this densest center and all other sampled features and plotted histograms to illustrate the distribution of these similarities. The histograms are provided in Fig.~\ref{histogram_train} and Fig.~\ref{histogram_test}. Additionally, we randomly visualized several patches corresponding to specific cosine similarity values.

As shown in Fig.~\ref{histogram_train} (a) (histogram for 32,768 patches from negative slides in the training set), approximately 92\% of patches have cosine similarities above 0.9 with the anchor, and 21,092 patches (64\%) even exceed 0.999. This clearly demonstrates that patch features from negative slides in the training set are tightly clustered under the Similarity Loss. In contrast, as shown in Fig.~\ref{histogram_train} (b) (histogram for 32,768 patches from positive slides in the training set), patch features from positive slides are more dispersed, with cosine similarities ranging from –0.5 to 1 under the SimCLR loss, and only 4,307 patches (13\%) falling between 0.9 and 1. Some randomly selected patches from positive slides with different cosine similarities shown in Fig.~\ref{histogram_train} (b) were also reviewed by our pathologists. Patches marked with red solid boundaries were diagnosed as “definitely/probably positive”, while those marked with red dashed boundaries were considered “possibly positive”. Among the randomly selected patch examples for visualization, we observed that patches with cosine similarities close to or below –0.2 were consistently diagnosed as positive, whereas patches with similarities close to or above 0.2 were consistently diagnosed as negative. Patches with cosine similarities near 0 were located in a transition zone, with some diagnosed as “possibly positive.” These diagnostic results clearly demonstrate that WeakSupCon training naturally separates negative and positive patches without requiring patch-level labels.

Fig.~\ref{histogram_test} presents the feature distributions in the Camelyon16 test set. In the training set, negative patches from negative slides were only directly optimized by the Similarity Loss, while negative patches from positive slides were only directly optimized by the SimCLR Loss. However, negative patches in the test set, no matter in negative slides or positive slides, are entirely unseen by the model and are therefore influenced by both the Similarity Loss and the SimCLR Loss in their feature distributions. Consequently, we observed different distribution patterns in the test set. Specifically, the feature distributions of patches from negative and positive slides in the test set became more similar, which is consistent with theoretical expectations. In the Camelyon16 dataset, more than 90\% of patches in positive slides are actually negative \cite{shao2021transmil, Li2020-da-dsmil}; these negative patches share the same labels, textures, and morphological patterns as patches from negative slides, leading to similar feature distributions. Nevertheless, as demonstrated by the patch examples and the corresponding pathologist diagnoses, patches from positive slides with lower cosine similarities to the anchor are still likely to be positive. Conversely, patches with cosine similarities close to or above 0.2 are likely to be negative.

\begin{table*}[t]
\centering
\caption{Ablation studies on the weight of Similarity Loss in WeakSupCon.
}\label{tab3_ablation_study}
\setlength{\tabcolsep}{3pt}
\begin{tabular}{|l|l|l|l|l|l|l|}
\hline
 & \multicolumn{3}{|c|}{DTFD-MIL Results on Camelyon16 dataset} & \multicolumn{3}{|c|}{AB-MIL Results on Camelyon16 dataset} \\ 
\hline
%  & \multicolumn{3}{|c|}{DTFD-MIL Results} & \multicolumn{3}{|c|}{AB-MIL Results} \\ 
% \hline
% \multicolumn{7}{|c|}{Camelyon16 dataset}\\ 
% \hline
\textbf{Weight} &  \textbf{Balanced acc} & \textbf{Accuracy} & \textbf{AUC} &  \textbf{Balanced acc} & \textbf{Accuracy} & \textbf{AUC}\\
\hline
0.25  &  0.8779$\pm$0.0083 & 0.8992$\pm$0.0078 & 0.8699$\pm$0.0176 &  0.8746$\pm$0.0297 & 0.8837$\pm$0.0135 & 0.9152$\pm$0.0682\\
\hline
1.0  &  {\bfseries 0.9265}$\pm$0.0036  & {\bfseries 0.9431}$\pm$0.0044 & {\bfseries 0.9694}$\pm$0.0018 &  {\bfseries 0.9318}$\pm$0.0042  & {\bfseries 0.9431}$\pm$0.0044 & {\bfseries 0.9699}$\pm$0.0009\\
\hline
4.0  &  0.5000$\pm$0.0000 & 0.6202$\pm$0.0000 & 0.5000$\pm$0.0000 &  0.5000$\pm$0.0000 & 0.6202$\pm$0.0000 & 0.5000$\pm$0.0000\\
\hline
 & \multicolumn{3}{|c|}{DTFD-MIL Results on RVT dataset} & \multicolumn{3}{|c|}{AB-MIL Results on RVT dataset} \\ 
\hline
% \multicolumn{7}{|c|}{Renal vein thrombosis (RVT) dataset}\\ 
% \hline
\textbf{Weight} &  \textbf{Balanced acc} & \textbf{Accuracy} & \textbf{AUC} &  \textbf{Balanced acc} & \textbf{Accuracy} & \textbf{AUC}\\
\hline
0.25  &  0.7769$\pm$0.0058 & 0.7816$\pm$0.0199 & 0.8485$\pm$0.0253 &  0.7744$\pm$0.0296 & {\bfseries 0.7931}$\pm$0.0345 & 0.8603$\pm$0.0077\\
\hline
1.0  &  {\bfseries 0.8014}$\pm$0.0130  & {\bfseries 0.8046}$\pm$0.0199 & {\bfseries 0.8771}$\pm$0.0177 &  {\bfseries 0.7980}$\pm$0.0307  & {\bfseries 0.7931}$\pm$0.0000 & {\bfseries 0.8788}$\pm$0.0315\\
\hline
4.0  &  0.5000$\pm$0.0000 & 0.6207$\pm$0.0000 & 0.5000$\pm$0.0000 &  0.5000$\pm$0.0000 & 0.6207$\pm$0.0000 & 0.5000$\pm$0.0000\\
\hline
 & \multicolumn{3}{|c|}{DTFD-MIL Results on kidney metastasis dataset} & \multicolumn{3}{|c|}{AB-MIL Results on kidney metastasis dataset} \\ 
\hline
%\multicolumn{7}{|c|}{Kidney metastasis dataset}\\ 
%\hline
\textbf{Weight} &  \textbf{Balanced acc} & \textbf{Accuracy} & \textbf{AUC} &  \textbf{Balanced acc} & \textbf{Accuracy} & \textbf{AUC}\\
\hline
0.25  &  {\bfseries 0.9091}$\pm$0.0152 & {\bfseries 0.8696}$\pm$0.0218 & {\bfseries 0.9293}$\pm$0.0027 &  {\bfseries 0.9141}$\pm$0.0175 & {\bfseries 0.8768}$\pm$0.0251 & {\bfseries 0.9293}$\pm$0.0014\\
\hline
1.0  &  {\bfseries 0.9091}$\pm$0.0000  & {\bfseries 0.8696}$\pm$0.0000 & 0.9277$\pm$0.0016 &  0.9040$\pm$0.0088  & 0.8623$\pm$0.0126 & 0.9246$\pm$0.0013\\
\hline
4.0  &  0.8858$\pm$0.0000 & {\bfseries 0.8696}$\pm$0.0000 & 0.9052$\pm$0.0000 &  0.8858$\pm$0.0000 & 0.8696$\pm$0.0000 & 0.8959$\pm$0.0013\\
\hline

\end{tabular}
\end{table*}

\subsection{Comparison with Foundation Models}
In multiple instance learning, foundation models are often used as encoders to generate patch features for downstream tasks. In our experiments, we also compared WeakSupCon with several foundation models, which is shown in Table~\ref{tab2_foundation_models}.

The ImageNet-pretrained encoder using the same backbone as WeakSupCon achieved the lowest performance in both DTFD-MIL and AB-MIL in most cases. This is likely because the encoder was pretrained on natural images, whereas our datasets consist of histopathology images, leading to a significant domain shift. This result underscores the importance of the feature quality for input into the MIL, as different encoders substantially affect MIL performance.

We further compared our model with state-of-the-art histopathology foundation models, including Prov-GigaPath \cite{xu2024whole} and UNI2-h \cite{chen2024uni}, which were pretrained on enormous histopathology datasets using much more powerful computational resources. In general, our WeakSupCon outperforms these histopathology foundation models on two out of the three datasets, including the renal vein thrombosis (RVT) dataset and kidney metastasis dataset. %The Camelyon16 dataset is a public dataset and is more likely to be in the training distribution of foundation models. In contrast, 
Notably, on the RVT dataset, WeakSupCon achieves the strongest performance, exceeding the foundation models by a substantial margin. Additionally, WeakSupCon demonstrates greater robustness across datasets, as it mitigates the challenge of domain shift by being trained directly on local data.

\subsection{Ablation Studies}
Since the Similarity Loss is a newly introduced loss term in our work, we conducted ablation studies to evaluate its influence by assigning different weights to it. As shown in Table~\ref{tab3_ablation_study}, we experimented with three different weight settings: 0.25, 1, and 4.

When the weight was increased to 4, we observed that the total loss did not decrease and remained unchanged during the early stage of training in both the Camelyon16 and RVT datasets. As a result, the encoder failed to learn meaningful representations and behaved like a random guesser. The most probable explanation is that the optimization was trapped in a local minimum. Without the SimCLR loss, the easiest shortcut for optimization on the Similarity Loss is to output identical features regardless of the input. By assigning a high weight, the training process overemphasizes the generation of similar features, thereby neglecting feature diversity.

On the other hand, when the weight of the Similarity Loss was reduced to 0.25, we also did not observe clear advantages. While we found slight improvements in the kidney metastasis dataset, more pronounced drawbacks were observed in the other two datasets, particularly Camelyon16. Theoretically, setting the weight to a low value results in weaker supervision based on bag-level labels, making the training process resemble SimCLR training applied only to patches from positive bags.

In comparison, setting the weight to 1 appears to be the most balanced and natural choice, as it assigns equal importance to each sample regardless of its bag-level label.

\section{Conclusion}

In our work, we proposed a novel feature representation learning method called WeakSupCon designed for multiple instance learning (MIL) settings where only bag-level labels are available. Compared to self-supervised learning methods, our approach effectively leverages bag-level labels to achieve a clearer separation between positive and negative instances in the latent feature space, even in the absence of instance-level annotations. This improved feature separation significantly enhances the performance of downstream MIL tasks. Moreover, comparisons with state-of-the-art histopathology foundation models further demonstrate the robustness and generalizability of our proposed method.

\section*{CRediT author statement}

\textbf{Bodong Zhang}: Conceptualization, Methodology, Software, Validation, Formal analysis, Investigation, Data Curation, Writing - Original Draft, Visualization. 

\textbf{Xiwen Li}: Software, Formal analysis. 

\textbf{Hamid Manoochehri}: Visualization. 

\textbf{Xiaoya Tang}: Visualization. 

\textbf{Deepika Sirohi}: Resources, Data Curation. 

\textbf{Beatrice S. Knudsen}: Resources, Data Curation, Writing - Review \& Editing, Supervision, Project administration, Funding acquisition. 

\textbf{Tolga Tasdizen}: Resources, Writing - Review \& Editing, Supervision, Project administration, Funding acquisition.

% \section*{Availability of research data/code}
% Our related research data and code is publicly available at \url{github.com/BzhangURU/Paper_WeakSupCon_for_MIL}

\section*{Declaration of competing interest}
The corresponding author, on behalf of all the authors of a submission, has disclosed all financial and personal relationships with other people or organizations that could inappropriately influence (bias) their work.

\section*{Availability of research data/code}
The Camelyon16 dataset is available at \url{camelyon16.grand-challenge.org/Data} 

Our related research code is publicly available at \url{github.com/BzhangURU/Paper_WeakSupCon_for_MIL}

\section*{Acknowledgments}
The project was funded by NIH/NCI 1R21CA277381, DoD HT94252410186, and Department of Veterans Affairs I01CS002622. We acknowledge the support of the Computational Oncology Research Initiative (CORI) at the Huntsman Cancer Institute, ARUP Laboratories, and the Department of Pathology at the University of Utah.

\section*{Declaration of generative AI and AI-assisted technologies in the manuscript preparation process}

During the preparation of this work the authors used ChatGPT in order to assist with grammar correction and improvement of wording in certain sentences for better readability. All sentences were originally written by the authors and not generated by AI. After using this tool/service, the authors reviewed and edited the content as needed and take full responsibility for the content of the published article.

{\small
\bibliographystyle{ieee_fullname}
\bibliography{refs.bib}
}
\end{document}